\documentclass[conference]{IEEEtran}
\IEEEoverridecommandlockouts

\usepackage[]{graphicx}
\usepackage{amssymb,amsmath,amsthm}
\newtheorem{definition}{Definition}

\usepackage[T1]{fontenc}
\usepackage[utf8]{inputenc}

\usepackage[bookmarks=false]{hyperref}
\hypersetup{
            pdfborder={0 0 0},
            breaklinks=true}

\usepackage{todonotes}

\def\BibTeX{{\rm B\kern-.05em{\sc i\kern-.025em b}\kern-.08em
    T\kern-.1667em\lower.7ex\hbox{E}\kern-.125emX}}
\begin{document}

\title{MULTIMEDIA SEARCH AND TEMPORAL REASONING}

\author{\IEEEauthorblockN{ Removed for double blind review}}

\author{\IEEEauthorblockN{Marcio Moreno$^1$,~Rodrigo Santos$^2$,~Wallas Santos$^3$,~Sandro Fiorini$^4$,~Reinaldo Silva$^5$,~Renato Cerqueira$^6$}
\IEEEauthorblockA{\textit{IBM Research Brazil} \\
Rio de Janeiro, Brazil \\
\{mmoreno$^1$,~rcerq$^6$\}@br.ibm.com \\
\{rodrigo.costa$^2$,~wallas.sousa$^3$,~sandro.fiorini$^4$\}@ibm.com \\
cmmozart@gmail.com$^5$
}
}

\maketitle

\begin{abstract} Properly modelling dynamic information that changes over time still is an open issue. Most modern knowledge bases are unable to represent relationships that are valid only during a given time interval.  In this work, we revisit a previous extension to the hyperknowledge framework to deal with temporal facts and propose a temporal query language and engine. We validate our proposal by discussing a qualitative analysis of the modelling of a real-world use case in the Oil \& Gas industry.
\end{abstract}


\section{Introduction}
Modern knowledge-based systems still have difficulties in dealing with information that change over time, known as \emph{dynamic information}. In several real-world scenarios, whatever phenomenon we represent (natural, computational or abstract) is unlikely to be static, therefore representing dynamic information is vital \cite{Fisher2008}. For instance, one can use any knowledge representation model to represent that a given person is the CEO of a company. However, this information no longer holds when another employee assumes that position. This simple example helps to illustrate the need for representing the time span in which a given information is valid.

RDF is a representation language widely used for specifying and querying information in knowledge bases. Although it has proven itself in a widespread range of scenarios, its \emph{subject-predicate-object} (SPO) data model lacks the proper expressiveness for representing temporal information about facts. Different approaches have been proposed over the years for addressing this issue, but their use in current knowledge-based systems is not widespread.

In this work, we approach the problem by using the \emph{hyperknowledge} model \cite{Moreno2017}. Hyperknowledge is a hybrid knowledge representation framework that unifies concepts from hypermedia and knowledge engineering. By using hyperknowledge, one can specify applications linking, for instance, knowledge descriptions and hypermedia anchors (segments of a media object) and  knowledge-aware interactions (e.g., a given content should be presented every time users interact with specific concepts).  Previous work \cite{Moreno2018} extended hyperknowledge with notion of temporal anchors (timed segments in media objects), which allowed the specification of temporal facts in the knowledge base.  In this paper, we revisit that temporal framework and implement it in a query language and engine that is capable of running temporal queries on temporal hyperknowledge models. We illustrate the practical gains by discussing how our approach can be used in a real-world scenario of the Oil \& Gas Industry.

\section{Related Work}

Temporal reasoning in knowledge bases is a well-known problem in AI . There are two key issues that a temporal reasoning framework should address \cite{Vila1994}: (i) an extension to the language/model for representing the temporal aspect of the knowledge, and (ii) a temporal reasoning system. In this paper, we focus on the first issue.

 A proper formalism should provide means to link atemporal assertions (e.g., facts) to temporal references \cite{Pani2001}.  Most of the current AI-based systems rely on the SPO-based RDF data model for representing information. However, representing facts as SPO triples complicates the addition of time information about validity.
 There are some works in literature that address such issue. One of the earliest attempts to add time to RDF was made by Buraga and Ciobanu \cite{Buraga2002}. The authors proposed an extension to the XML representation of RDF triples to add temporal relationships (called \emph{links}) to entities. Their proposal allowed one to express all the Allen operators \cite{Allen1990} among web pages. Despite having a specific scope (expressing timing relationships between websites) and limited expressiveness (it does not promote the specification that a fact is valid over a time period) it helps to illustrate the lack of time awareness in RDF-based languages.

In a well-known work,  \cite{Gutierrez2005} proposes the Temporal RDF model, which extends the SPO data model by labeling triples with temporal values. Such temporal values represent the time in which a triple is valid. This proposal effectively redresses the triple model as a quad model. While being an interesting approach, it does not allow for the explicit specification of multiple validy intervals for a given triple (except for triple reification, which is not ideal in terms of storage and retrieval capacity). A more recent and similar approach to Temporal RDF is presented in \cite{Bereta2013} and but has similar drawbacks.

In \cite{Hoffart2013}, Hoffart \emph{et al.} present YAGO2, an extension of the YAGO knowledge-base that supports the specification of spatiotemporal relationships. Their approach is also based on RDF, using reification to temporaly qualify triples. As mentioned before, such approaches are not ideal in terms of storage and retrieval performance.


There are other works in literature either based on those discussed in this section \cite{Rula2014,Sheth2008} or that extend SPARQL to allow temporal queries in knowledge-bases based on the RDF data model \cite{Lopes2010,Tappolet2009,Perry2011}. RDF lacks enough expressiveness for expressing $n$-ary relationships, which hinders the direct specification of temporal information, leading to several proposals using different techniques to overcome this issue.

Our proposal is built on the hyperknowledge model. Hyperknowledge was proposed in \cite{Moreno2017} as a hybrid knowledge representation model. Hyperknowledge  supports the representationof $n$-ary relationships without reification, which allows one to express using a single relationship all temporal intervals in which a fact holds. This facilitates the implementation of reasoning engines and also avoids the duplication of facts for adding a temporal dimension to them.

\section{Motivational Scenario}
\label{motivational-scenario}

In this section, we present a use case in the Oil \& Gas industry that illustrates the need for representing temporal information in knowledge bases. We will refer to this sceario thoroughout the next sections.

\begin{figure}
	\centering
	\includegraphics[width=0.8\linewidth]{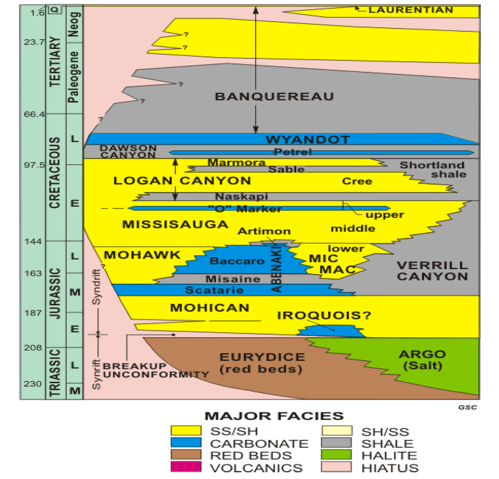}
	\caption{Chronostratigraphic chart of the Scotia Basin, offshore Nova Scotia, Canada \cite{Wade1990} (\emph{Neog} corresponds to Neogene and \emph{E, M, L} abreviate Early, Middle and Late, respectively)}
	\label{fig:chrono}
\end{figure}

A common task in Petroleum Geology is the characterization of basin formation. Geologists must characterize the temporal evolution of  geological formations in a given basin throughout geological time in order to draw conclusions regarding whether it is possible to have a proper condition for hydrocarbon accumulation and/or production. As an illustration, consider Fig.~\ref{fig:chrono}. It depicts a chronostratigraphic chart of the Scotia Basin, located offshore of Nova Scotia, Canada. This type of chart corelates geological periods with litho-stratigraphic units (here called ``facies'') formed on those periods. The first three columns represent geological time in different scales (absolute time in millions of years, periods and epochs, respectively). The fourth column shows information about geological formations/facies and lithology, where each color represents a different formation.

Based on this type of data, geologists can make temporally-situated queries based on temporal relations about the formation of those facies, such as ``\emph{which facies are forming in which geological periods in a given basin}'', or ``\emph{what are the facies of a given type forming after a given epoch}''. These queries require a representation of geological facts that are valid only in a given (geological) timeframe.  They require temporal reasoning in order to calculate temporal relationships between facts.



\section{Preliminaries: The Hyperknowledge Model}

\begin{figure}
	\centering
	\includegraphics[width=\linewidth]{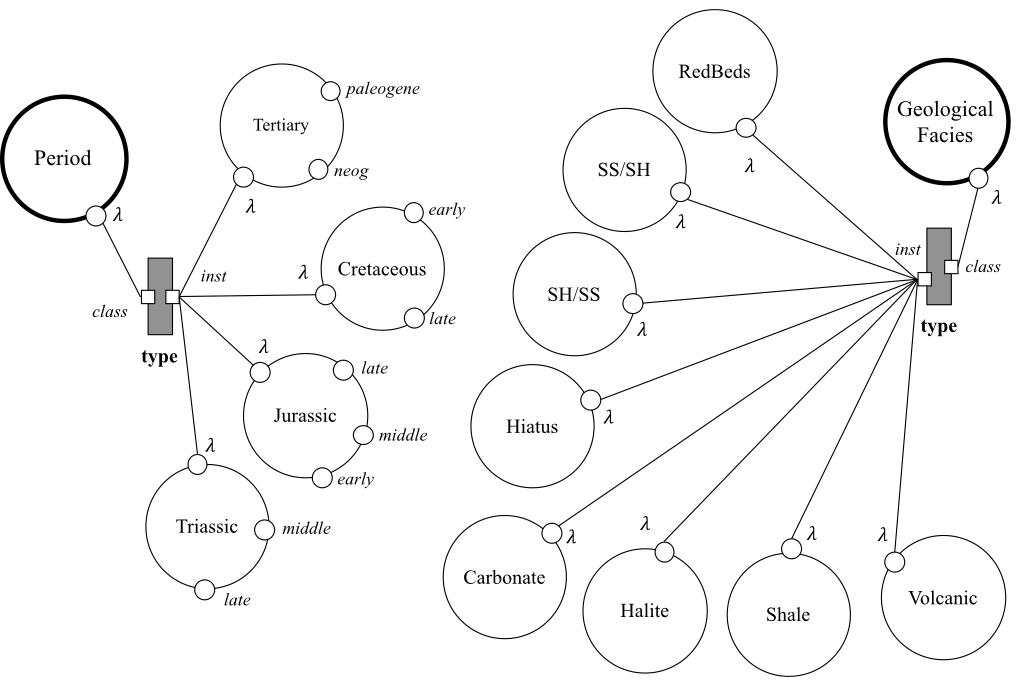}
	\caption{Ontology using the hyperknowledge model.}
	\label{fig:onto}
\end{figure}

Hyperknowledge was first proposed aiming to better represent relationships among multimedia content (e.g., image, audio, video, text, etc.) as well as conceptual entities, allowing the specification of multimedia-aware knowledge bases. As a first example, consider the model in Fig.~\ref{fig:onto}. It illustrates taxonomic information about geological facies and geolgoical periods present in Fig.~\ref{fig:chrono} as hyperknowledge entities.

A hyperknowledge data model can be represented as a graph with enhanced vertices. In Fig.~\ref{fig:onto}, there are two types of vertices. Circular labeled vertices are \emph{nodes}, which are entities that represent media content or abstract concepts. For instance, the node labeled \emph{Period} represents the chronological concept of geological periods. Likewise, the node labeled \emph{Geological Facies} represents the homonym abstract concept.

Hyperknowledge nodes are decorated by \emph{anchors}, depicted as small labeled circles on the node boundaries. Anchors are first-class entities that represent (i.e. select) a portion of the node's content. A spatial anchor may represent a subregion of an image. A temporal anchor represents a temporal segment (interval) of a continuous media (e.g., audio, video). In Fig.~\ref{fig:onto}, portions of geological periods that correspond to geological epochs are depicted as labeled anchors. The \emph{lambda} ($\lambda$) anchors represent the whole content of the node.

Darkgrey vertices in Fig.~\ref{fig:onto} are \emph{links}, which define $n$-ary relationships among nodes. One of links' main characteristics is that they can associate nodes only through their anchors.  Links have a \emph{type}, a \emph{predicate}  and some \emph{roles}.  The predicate of a link defines its name. Roles specify the different arguments of the relationship represented by the link. A link type defines its overall role structure. In this paper, we employ \emph{fact} and \emph{hierarchy} links only. Fact links define facts having a subject and an object, similar to SPO triples, but they can also express $n$-ary relationships in hyperknowledge. Hierarchy links are used to define instantiation relationships among nodes. In our scenario, the link named \emph{type} represents a $n$-ary relation between a class and $n-1$ instances of that class.

\section{Representing Temporal Relationships in Hyperknowledge}

In \cite{Moreno2018}, the hyperknowledge model has been extended with a temporal model to represent temporal information. In this section, we revisit that temporal model, providing clearer definitions to some of its main concepts. In the following sections, we employ these notions to propose a temporal query engine.

The main idea of the temporal extension to hyperknowledge is to exploit the association of links and nodes through anchors in order to represent complex temporal relations. A temporal anchor is defined by the hyperknowledge model as being a temporal segment of a node representing a (continuous) media content. For instance, one can define a node that represents a video and create a temporal anchor that represents an interval of that video.

All anchors are assumed to have timing information, including concept nodes. This facilitates the implementation of the reasoning engine (described in next section) because it does not have to handle media nodes differently from others due to their temporal dimension. In fact, any anchor defines a temporal interval and may be used in the temporal algebra implemented by the reasoning system.

We start by defining time. We will consider time to be: (a) \emph{Interval-based} rather then point-based intervals; (b) \emph{unbounded} in the past and future; and (c) \emph{linear}, with no notion of branching, parallelism or circularity. We assume closed time intervals, where $[a, b]$ denotes a closed interval starting at time $a$ and finishing at time $b$. The only exception is with unbounded time intervals $[-\infty,b], [a,+\infty], [-\infty,+\infty]$, in which we assume to be open on the unbounded side.

%
%


\begin{definition}
	A \emph{temporal} anchor is an anchor with a ---possibly unbounded --- time interval $[begin, end]$.
\end{definition}

We also refer to a time interval $[begin, end]$ of an anchor $a$ by using the dot notation; i.e.  $a.begin$ and $a.end$.  The interpretation of a temporal anchor is given by its node. For example, a temporal anchor might denote a temporal slice in a piece of media stream or a part of the temporal existence of a given entity.

Temporal anchors can be temporally included in each other:

\begin{definition} Let $x$ and $y$ be two temporal anchors. Anchor $x$ is said to be \emph{included} in $y$ (denoted $in(x, y)$ iff $ y.begin \leq x.begin$ and $x.end \leq y.end$.
\end{definition}

This notion allows us to define temporal lambda anchors. Let $A(n)$ denote the set of anchors of a node $n$. Then::

\begin{definition}A temporal $\lambda$ anchor represents the whole time interval on which a node is defined, such that, given any node $n$, its temporal lambda anchor $\lambda$ and any of its anchors $a \in A(n)$, then $in(n.a,n.\lambda)$.
\end{definition}




An example of temporal lambda anchors are anchors denoting the whole time of a piece of stream, as well as, denoting an entity existence interval.

\begin{figure}
\centering
\includegraphics[width=0.9\linewidth]{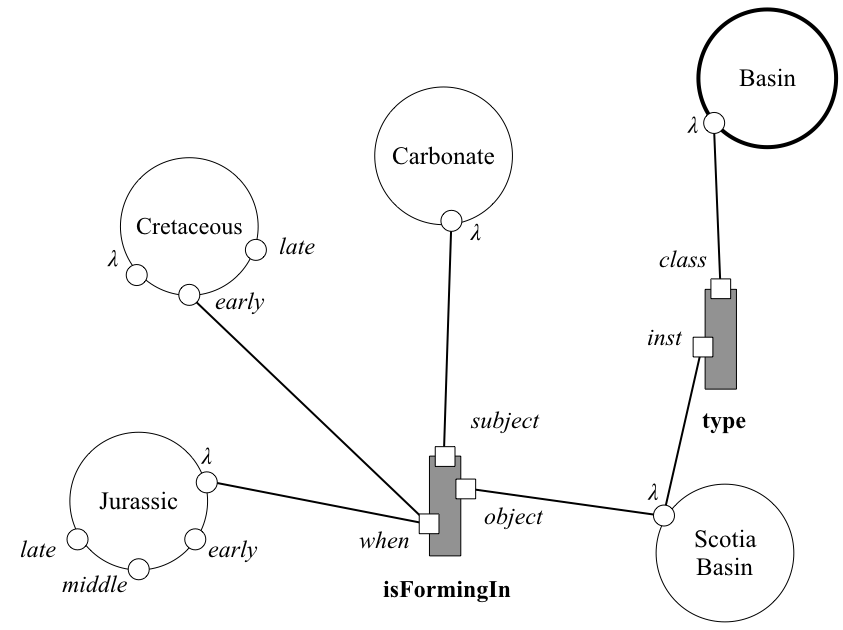}
\caption{Hyperknowledge modelling representing S1 and S2.}
\label{fig:models1s2}
\end{figure}

Reconsider Fig.~\ref{fig:onto}. One can define that the anchors \emph{late}, \emph{middle} and \emph{early} in the \emph{Jurassic} concept node represent time intervals within the Jurassic period. These can be used to represent temporal relationships with facies related to their formation. Fig.~\ref{fig:models1s2} depicts a hyperknowledge specification of the fact that the Scotia Basin has a Carbonate facies that is being formed along the whole of Jurassic Period and early Cretaceus. This fact is modulated by the \emph{when} role, which states that this relationship is true only during the Jurassic Period and the early Cretaceous.

%

We introduce the role \emph{when} in order to differentiate static and dynamic facts: the former represents aspects of the world that never changes; the latter expresses the dynamic aspect representing an information that may change over the time. Informally, the role \emph{when} defines a set of intervals in which the relationship expressed by a link should be considered valid (or \emph{true}):

\begin{definition} Let $l$ be a link and $W(l)$ be the set of all temporal anchors bound to the \emph{when} role of $l$. The link $l$ is considered \emph{temporally valid} in the interval $[x,y]$ if, for any $t$, such that $x \leq t \leq y$, there is a temporal anchor $a \in W(l)$ such that $a.begin  \leq t \leq a.end$. If $W(l) = \emptyset$, then $l$ is considered atemporally valid (i.e. temporaly valid within $[-\infty,+\infty]$).
\label{def:hklink}
\end{definition}
%
Note that this definition does not change the semantics of links that do not use the role \emph{when}. For instance, consider the link defining that \emph{Scotia Basin} is an instance of a \emph{Basin.} As this link has no role \emph{when} (i.e., $W(l) = \emptyset$), it is considered valid for any interval. That is, the hierarchy relationship expressed by that link always holds.

\section{Queries}
\label{sec:queries}

In this section, we describe a query language to retrieve information from a knowledge base structured using hyperknowledge.
The queries have the general form ``\emph{\textbf{select} targets \textbf{where} constraints}''. The targets define the type of the output of the query. The possible types are nodes, properties, or anchors.

Query variables refer to \emph{instances} and \emph{classes} of the hyperknowledge model. The constraints are  links, temporal operators, and attribute comparisons. The link constraint has the form ``\emph{predicate(role\_1:a role\_2:b ... role\_n:n)}'', allowing querying $n$-ary relations. Role can be omited if they are not used in constraints. The temporal operators are equivalent to those defined by Allen \cite{Allen1990}, e.g., ``\emph{A before B}''. Finally, constraints may compare the value of a nodes' property with other property or literals.

%
%
%
%
%
%

In the query evaluation process, the reasoning engine must resolve the type hierarchy. In practice, the goal of the engine is to find all instances that have relations and properties that satisfy the query constraints. Therefore, when a node is tested against a constraint, the reasoning engine should check if it is \emph{true} by replacing each instance of that type considering the full type hierarchy.

In our current proposal, there are three types of constraints:

\begin{itemize}
\item
  \emph{Link Constraint:} A link constraint is evaluated to \emph{true}, if a link exists with the given predicate and for each declared role there are nodes that match with the passed identifier, either as an instance or their classes. If the link has a constraint based on a time interval, temporal validty is checked (according with Def. \ref{def:hklink}).
\item
  \emph{Attribute constraint:} It checks if the attribute in an instance (property) satisfies a comparison with the literal of the constraint;
\item
  \emph{Temporal constraints:} Constraint like ``\emph{foo before bar}'', which are evaluated by the reasoning engine by checking if at least one of its anchors satisfies the constraint.
\end{itemize}



Our query language can handle common queries in our domain. For example, fetching all the basins in which carbonates are forming:
\begin{equation}
\begin{aligned}
& \text{\textbf{select }}  \text{{Basin} \textbf{ where }} \\[-4pt]
& \quad \text{\emph{{isFormingIn}}(\emph{subject}:{Carbonate} \emph{object}:Basin)}
\end{aligned}
\end{equation}
This query has only a simple link constraint. It retrieves instances of the node \emph{Basin} that in their links there is a role bound to the instance \emph{Carbonate}. Temporal queries can be specified with the operator \emph{when}, such as retrieving the basin having a carbonate facies forming after the Jurassic period:
\begin{equation}
\label{eqn:2}
\begin{aligned}
& \text{\textbf{select }}  \text{{Basin} \textbf{ where }} \\[-4pt]
	& \quad \text{\emph{{isFormingIn}}(\emph{subject}:{Carbonate} \emph{object}:Basin}\\[-4pt]
	& \quad \quad \text{ \emph{when}:{Period})} \\[-4pt]
	& \quad \text{\textbf{and} {Period} \textbf{after} Jurassic}
\end{aligned}
\end{equation}
In this example, we combine two constraints, the same from previous query and the temporal operation \emph{after}. Here we have used the role \emph{when} and declared the variable \emph{Period} implicitly. Then, we compared \emph{Period} with the interval defined by the \emph{lambda} anchor of the \emph{Jurassic} instance. Anchors can also be refered in the queries, such as when selecting the periods after the Jurassic in wich a carbonate facies has been identified being in formation.
\begin{equation}
\begin{aligned}
&\text{\textbf{select }} \text{\textbf{anchor from} {Period} \textbf{ where }} \\[-4pt]
& \quad \text{\emph{{isFormingIn}}(\emph{subject}:{Carbonate} \emph{object}:Basin}\\[-4pt]
& \quad \quad \text{ \emph{when}:{Period})} \\[-4pt]
& \quad \text{\textbf{and} {Period} \textbf{after} Jurassic}
\end{aligned}
\end{equation}



\section{Implementation}

\begin{figure}
\centering
\includegraphics[width=0.9\linewidth]{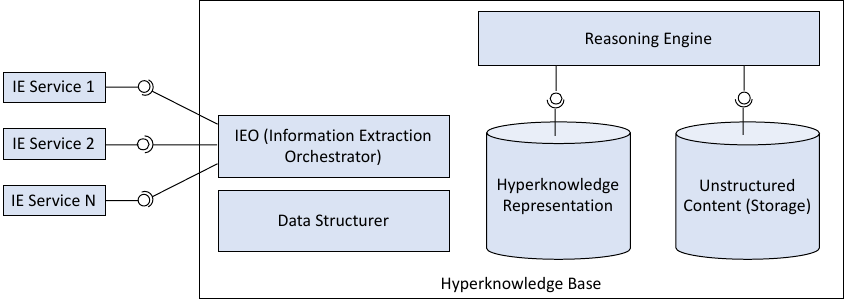}
\caption{The Hyperknowledge Base architecture overview.}
\label{fig:architecture}
\end{figure}

We have implemented a knowledge-base system (called Hyperknowledge base, or HKBase{} as shorthand) that uses hyperknowledge as data model. Fig.~\ref{fig:architecture} depicts the main components of the HKBase{} architecture.

The HKBase{} implements a CRUD API for knowledge curation and has functionalities for handling multimedia data. The architecture includes two storages, one to store the hyperknowledge content, and a second one to store unstructured content, such as multimedia data that need to be uploaded to the system.  The \emph{Information Extraction Orchestrator} (IEO) is a component that selects an apropriate information extraction service to get concepts from every new media content referenced by the hyperknowledge representation. For instance, if a hyperknowledge node represents an image having an URL property pointing to a valid and accessible file, the HKBase{} is capable of retrieving that image, and to choose an information extraction service (if available) specialized in processing that type of content to extract semantic information from it. The \emph{data structurer} receives the output from IEO and structures it before saving it to the hyperknowledge base. Finally, a reasoning engine processes queries made by the user.

\subsection{Mapping Hyperknowledge to a Knowledge Graph}

The HKBase{} does not rely on a specific storage service, but rather, it defines a driver API (called \emph{IDB -- Interface for DataBase}) which should be implemented by any given storage to be used. The main functionality of an IDB driver is to map the hyperknowledge representation maintained by the HKBase{} to the data model supported by the underlying database. We currently have the Janus Graph\footnote{http://janusgraph.org/} implementation that fully supports the temporal reasoning.

The Janus Graph driver uses the Apache TinkerPop framework\footnote{http://tinkerpop.apache.org/} to convert the hyperknowledge entities to a knowledge graph representation supported. To implement this mapping, it uses different types of vertices and edges. A hyperknowledge node is represented by a vertex named \emph{node} and an anchor of hyperknowledge nodes is represented by an \emph{anchor} vertex. Anchor vertices hold in an internal table the properties of the corresponding hyperknowledge anchor.

Hyperknowledge links are represented by a third type of vertex called \emph{link}, which holds the predicate and roles in its properties table. This design allows representation of $n$-ary relationships, as those discussed above. This addresses the issue of edges having the limitation of connecting exactly two vertices in a graph.

Likewise, different types of edges are used to represent the relationships among vertices in the graph. A \emph{hasAnchor} edge connects a node to its anchors. A \emph{bind} edge connects an \emph{anchor} to a \emph{link} vertice. Bind edges specialize in its properties table to which link role that anchor is bound.

\begin{figure}
\centering
\includegraphics[width=\linewidth]{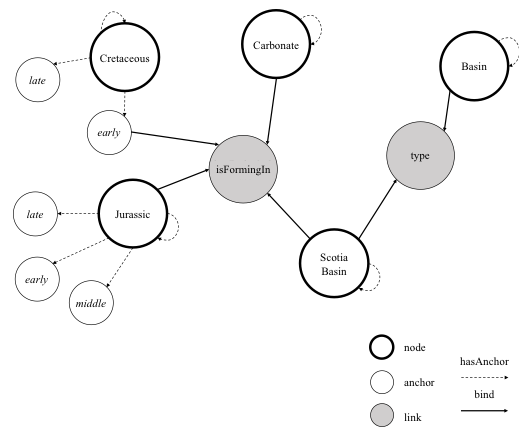}
\caption{Example of representation of the hyperknowledge model in a Graph Database.}
\label{fig:example}
\end{figure}

Fig.~\ref{fig:example} depicts the knowledge graph that represents the hyperknowledge model in Fig.~\ref{fig:models1s2}. Arrows represent edges and circles represent vertices in the figure.

The full conversion of all hyperknowledge entities to a knowledge graph requires additional types of vertices (\emph{context, switch, connector, etc.}) and edges (\emph{hasConnector, hasChildren, etc).} However, it is out of the scope of this paper the description of the whole process, but we have focused on the entities involved in the temporal reasoning.

\subsection{Temporal Reasoning}

The reasoning engine receives a query as input, processes it, and returns a set corresponding to the results of the query. For the case of temporal queries, the reasoning process is implemented in a two-stages pipeline composed of the following steps: i) data retrieval; and ii) computation of temporal relations and constraints. 


Consider Query \ref{eqn:2} in Section \ref{sec:queries}. In the first stage, the reasoning engine retrieves all links that have the predicate \emph{isFormingIn}, the role \emph{subject} bound to an instance of the class \emph{Carbonate}, the role \emph{object} bound to the node \emph{Basin}, and the role \emph{when} bound to an instance of \emph{Period}. The IDB Driver implements this retrieval by using the Gremlin query language\footnote{https://tinkerpop.apache.org/gremlin.html}.

The output of the first stage becomes the input to the second, which is entirely performed by the reasoning engine. For Query \ref{eqn:2}, it computes which of the returned periods satisfy the temporal constraint \emph{``after Jurassic''}.

The main advantage of this two-stages pipeline for processing queries is that the reasoning process becomes independent of storage services. That is, if all temporal calculus would be performed by the underlying database, the architecture of the HKBase{} would be dependent of that service. Using this two-stages processing, the IDB driver is responsible only for retrieving data from the database, which is an operation supported by any storage service.

\section{Final Remarks}

In this paper, we have revisited an approach for representing temporal information in hyperknowledge representation language. Temporal anchors provide a useful mechanism for expressing richer temporal relationships in hypermedia knowledge bases, being suitable for expressing the intervals in which a given fact holds. We also introduce a query language and engine that can take advantage of these constructs for temporal queries. As future work, we plan to finish the implementation of other IDB drivers and report a quantitative comparison analyses of our reasoning engine working with different storage services. We believe that such quantitative analysis can bring valuable insights to the Multimedia community.  We also intend to conduct user experiments regarding the temporal reasoning and report in a future work the qualitative analysis of users' perspective about the work described in this paper.



\bibliographystyle{IEEEbib}
\bibliography{paper.bib}

\end{document}